\documentclass[conference,a4paper]{APSIPA2017}
\usepackage{multirow}
\usepackage{amsmath}
\usepackage[psamsfonts]{amssymb}
\usepackage{amsxtra}
\usepackage{graphicx,color}
\usepackage{threeparttable}
\usepackage{url}
\usepackage{epsfig,epstopdf}

\begin{document}

\title{AP17-OLR Challenge: Data, Plan, and Baseline}

\author{%
\authorblockN{%
Zhiyuan Tang\authorrefmark{1},
Dong Wang\authorrefmark{1},
Yixiang Chen\authorrefmark{1} and
Qing Chen\authorrefmark{2}
}
\authorblockA{%
\authorrefmark{1}
Center for Speech and Language Technologies, Tsinghua University\\
Corresponding author: wangdong99@mails.tsinghua.edu.cn}
\authorblockA{%
\authorrefmark{2}
SpeechOcean\\
E-mail: chenqing@speechocean.com}
}

\maketitle
\thispagestyle{empty}

\begin{abstract}

We present the data profile and the evaluation plan of the second oriental language recognition (OLR) challenge AP17-OLR.
Compared to the event last year (AP16-OLR), the new challenge involves more languages and focuses more on
short utterances. The data is
offered by SpeechOcean and the NSFC M2ASR project. Two types of baselines are constructed to assist the participants,
one is based on the i-vector model and the other is based on various neural networks.
We report the baseline results evaluated with various metrics defined by the AP17-OLR evaluation plan
and demonstrate that the combined database is a reasonable data resource for multilingual research.
All the data is free for participants, and the Kaldi recipes for the baselines have been published online.

\end{abstract}

\section{Introduction}

Oriental languages can be divided into several language families,
including  Austroasiatic languages (e.g.,Vietnamese, Cambodia )~\cite{sidwell201114},
Tai-Kadai languages (e.g., Thai, Lao), Hmong-Mien languages (e.g., some dialects in south China), Sino-Tibetan languages (e.g., Chinese Mandarin), Altaic languages (e.g., Korea, Japanese), Indo-European languages (e.g., Russian)~\cite{ramsey1987languages,shibatani1990languages,comrie1996russian}.
The increasing demographic migration and international business interaction
have caused rich multilingual phenomena, e.g., codes switching from a primary language to a second language, and then
switching back, where the second language part is just one or two words.
Due to the geographical proximity, oriental languages influence each other,
resulting in complex development patterns in terms of both phonetics and linguistics.
These complex patterns make the multilingual research on these languages
particularly attractive.

To meet the request of this research, the center for speech and language technologies (CSLT) at
Tsinghua University and SpeechOcean organized the
first oriental language recognition (OLR) challenge during APSIPA ASC 2016~\cite{wang2016ap16}.
The goal of the challenge was to demonstrate how the state-of-art language identification (LID) techniques
can discriminate oriental languages, as well as fostering new technologies in this direction.
This challenge was very successful: 8 teams from China mainland, Taiwan,
Singapore, and Germany submitted 9 submissions, and the best system submitted by the NUS-I2R joint team
achieved very good performance ($C_{avg}$=$0.0113$, EER\%=$1.09$). This result partly demonstrated that
with the present technology, it is possible to discriminate the typical oriental languages from
each other, at least under the conditions set by the challenge.
More details can be found from the challenge web site.\footnote{http://olr.cslt.org}

Encouraged by the success of AP16-OLR, we propose the second OLR challenge,
based on APSIPA ASC 2017. This new challenge, denoted by AP17-OLR, will involve more languages
and thus be more challenging. Besides the $7$ languages in AP16-OLR, $3$ new oriental
languages are involved: Uyghur, Kazakh and Tibetan. Uyghur and Kazakh belong to the Turkic language,
and Tibetan belongs to the Sino-Tibetan language. Both are major minority languages
in China and are spoken by relatively large populations. The data is offered by the NSFC M2ASR project\footnote{http://m2asr.cslt.org},
following the M2ASR Free Data Program.

Another feature of the new challenge is that we put more emphasis on short utterances. Since the
 performance on long utterances had been demonstrated in AP16-OLR, we start to tackle
the more challenging scenario with short utterances. More importantly, LID on short
utterances can find significant practical values, for example in speech recognition with code-switching.
Unfortunately, the pervious state-of-the-art i-vector model is not very suitable in the short-utterance scenario,
as it heavily relies on the distributional patterns of the acoustic features,
for which a long utterance is necessary. Recently, researchers found that
methods based on deep neural models potentially solve this problem. This neural approach utilizes
 various forms of deep neural networks (DNNs) to discover language sensitive features from short-term segments ~\cite{lopez2014automatic,gonzalez2014automatic,gelly2016divide,zazo2016language,lozano2015end,jin2016lid,kotov2016language,garcia2016stacked}.
To meet the practical request and reflect the recent research advantage, AP17-OLR moves its focus to
short utterances, e.g., 1 second or 3 seconds.

In the rest of the paper, we will present the data profile and the evaluation plan of the AP17-OLR challenge. To
assist participants to build their own submissions, two types of baseline systems are constructed, based on the
i-vector model and various DNN models respectively. The Kaldi recipes of these baselines can be downloaded from the
challenge web site.

\begin{table*}[htb]
\begin{center}
\caption{AP17-OL3 and AP16-OL7 Data Profile}
\label{tab:ol10}
\begin{tabular}{|l|l|c|c|c|c|c|c|c|}
 \hline
\multicolumn{3}{|c|}{\textbf{AP17-OL3}} & \multicolumn{3}{c|}{AP17-OL3-train/dev}  & \multicolumn{3}{c|}{AP17-OL3-test}\\
\hline
Code & Description & Channel & No. of Speakers & Utt./Spk. & Total Utt. & No. of Speakers & Utt./Spk. & Total Utt. \\
\hline
ka-cn & Kazakh in China & Mobile & 86 & 50  & 4200 &      86 &  20  & 1800 \\
\hline
ti-cn & Tibetan in China & Mobile & 34 & 330   & 11100 &    34 & 50  & 1800 \\
\hline
uy-id & Uyghur in China &  Mobile & 353 & 20   & 5800 &    353 & 5  & 1800 \\
 \hline
\hline
\multicolumn{3}{|c|}{\textbf{AP16-OL7}} & \multicolumn{3}{c|}{AP16-OL7-train/dev}  & \multicolumn{3}{c|}{AP16-OL7-test}\\
\hline
Code & Description & Channel & No. of Speakers & Utt./Spk. & Total Utt. & No. of Speakers & Utt./Spk. & Total Utt. \\
\hline
ct-cn & Cantonese in China Mainland and Hongkong & Mobile & 24 & 320 & 7559 & 6 & 300 & 1800 \\
\hline
zh-cn & Mandarin in China & Mobile & 24 & 300 & 7198        & 6 & 300 & 1800 \\
\hline
id-id & Indonesian in Indonesia &  Mobile & 24 & 320 & 7671 & 6 & 300 & 1800 \\
\hline
ja-jp & Japanese in Japan & Mobile & 24 & 320 & 7662        & 6 & 300 & 1800 \\
\hline
ru-ru & Russian in Russia & Mobile & 24 & 300 & 7190        & 6 & 300 & 1800 \\
\hline
ko-kr & Korean in Korea & Mobile & 24 & 300 & 7196          & 6 & 300 & 1800 \\
\hline
vi-vn & Vietnamese in Vietnam & Mobile & 24 & 300 & 7200    & 6 & 300 & 1800 \\
 \hline
\end{tabular}
\begin{tablenotes}
\item[a] Male and Female speakers are balanced.
\item[b] The number of total utterances might be slightly smaller than expected, due to the quality check.
\end{tablenotes}
\end{center}
\end{table*}

\section{Database profile}

Participants of AP17-OLR can request the following datasets for system construction.

\begin{itemize}
\item AP16-OL7: The standard database for AP16-OLR, including AP16-OL7-train, AP16-OL7-dev, and AP16-OL7-test.
\item AP17-OL3: A new dataset provided by the M2ASR project, involving three new languages. It contains AP17-OL3-train and AP17-OL3-dev.
\item THCHS30:  The THCHS30 database (plus the accompanied resources) published by CSLT, Tsinghua University~\cite{wang2015thchs}.
\end{itemize}

Participants can use all the above data to train their submission systems, but are required to report the results on
the development set. The development set includes AP16-OL7-test and AP17-OL3-dev. Note that the development data should be
excluded from the training data when constructing the pre-submission system and computing the dev results.

Besides the speech signals, the AP17-OL3 and AP16-OL7 databases also provide lexicons of all the 10 languages, as well
as the transcriptions of all the training utterances. These resources allow training acoustic-based or phonetic-based
language recognition systems. Training phone-based speech recognition systems is also possible, though
large vocabulary recognition systems are not well supported, due to the lack of large-scale language models.

A test dataset will be provided at the date of result submission. This test set involves two parts: AP17-OL3-test
and AP17-OL7-test. The latter involves 7 languages that is the same as in AP16-OL7. The details of these databases
are described as follows.

\subsection{AP16-OL7}

The AP16-OL7 database was originally created by Speechocean targeting for various speech processing tasks.
It was provided as the standard training and test data in the AP16-OLR challenge.
The entire database involves 7 datasets, each in a particular language. The seven languages are:
Mandarin, Cantonese, Indoesian, Japanese, Russian, Korean and Vietnamese.
The data volume for each language is about $10$ hours of speech signals recorded in
reading style. The signals were
recorded by mobile phones, with a sampling rate of $16$kHz  and a sample size of $16$ bits.

For Mandarin, Cantonese, Vietnamese and Indonesia, the recording was conducted in a quiet environment.
As for Russian, Korean and Japanese, there are $2$ recording sessions for each speaker: the first session
was recorded in a quiet environment and the second was recorded in a noisy environment.
The basic information of the AP16-OL7 database is presented in Table~\ref{tab:ol10},
and the details of the database can refer to the AP16-OLR challenge web site or the description paper~\cite{wang2016ap16}.

\subsection{AP17-OL7-test}

The AP17-OL7 database is a new dataset provided by SpeechOcean. This dataset contains 7 languages as in AP16-OL7,
each containing $1800$ utterances. The recording conditions are the same as AP16-OL7. This database is used as
part of the test set for the AP17-OLR challenge.

\subsection{AP17-OL3}

The AP17-OL3 database contains 3 languages: Kazakh, Tibetan and Uyghur, all are minority languages in China.
This database is part of the Multilingual Minorlingual Automatic Speech Recognition (M2ASR), which is
supported by the National Natural Science Foundation of China (NSFC). The project is a three-party collaboration, including Tsinghua University,
the Northwest National University, and Xinjiang University. The aim of this project is to construct speech recognition systems for five minor languages in China (Kazakh, Kirgiz, Mongolia, Tibetan and Uyghur). However, our ambition is beyond that scope: we hope
to construct a full set of linguistic and speech resources and tools for the five languages, and make them open and free for
research purposes. We call this the M2ASR Free Data Program. All the data resources, including the tools published in this paper, are released on the web site of the project.

The sentences of each language in AP17-OL3 are randomly selected from the original M2ASR corpus.
The data volume for each language in AP17-OL3 is about $10$ hours of speech signals
recorded in reading style.
The signals were recorded by mobile phones,
with a sampling rate of $16$ kHz and a sample size of $16$ bits.
We selected $1800$ utterances for each language as the development set (AP17-OL3-dev), and the rest is used as the
training set (AP17-OL3-train). The test set of each language involves $1800$ utterances, and is provided separately
and denoted by AP17-OL3-test.
Compared to AP16-OL7, AP17-OL3 contains much more variations in terms of recording conditions and
the number of speakers, which may inevitably  increase the difficulty of the challenge task.
The information of the AP17-OL3 database is summarized in Table~\ref{tab:ol10}.

\section{AP17-OLR challenge}

Based on the experience of AP16-OLR challenge, we call the AP17-OLR challenge.\footnote{\url{http://cslt.riit.tsinghua.edu.cn/mediawiki/index.php/OLR_Challenge_2017}}
Following the definition of NIST LRE15~\cite{lre15}, the task of the challenge is defined
as follows: Given  a  segment  of  speech  and  a  language  hypothesis (i.e.,  a  target
language  of  interest  to  be  detected),  the  task  is  to decide  whether  that
target  language  was  in  fact  spoken  in  the given segment (yes or no), based on an
automated analysis of the data contained in the segment.
The evaluation plan mostly follows the principles of NIST LRE15. It focuses on
the close-set condition, i.e., the language identification task.

The challenge focuses on short utterances. We define three test conditions according to the
length of the test utterances: 1 second condition, 3 second condition and full-utterance condition.
The test utterances of the 1 second condition and 3 second condition are randomly excerpted from the original
ones. If a test utterance is not sufficient long for the excerption, it is simply
discarded. The evaluation details are described as follows.

\subsection{System input/output}

The input to the LID system is a set of speech segments in unknown languages (but within
the $10$ languages of AP17-OL3 and AP16-OL7). The task of the LID system is to determine
the confidence that a language is contained in a speech segment. More specifically,
for each speech segment, the LID system outputs a score vector $<\ell_1, \ell_2, ..., \ell_{10}>$,
where $\ell_i$ represents the confidence that language $i$ is spoken in the speech segment.
Each score $\ell_i$ will be interpreted as follows: if $\ell_i \ge 0$, then the decision
would be that language $i$ is contained in the segment, otherwise it is not. The scores
should be comparable across languages and segments.
This is consistent with
the principle of LRE15, but differs from that of LRE09~\cite{lre09} where an explicit decision
is required for each trial.

In summary, the output of an OLR submission will be a text file, where each line contains
a speech segment plus a score vector for this segment, e.g.,

\vspace{0.5cm}
\begin{tabular}{ccccccccc}
seg$_1$ & 0.5  & -0.2 & -0.3 &  ... & 0.1 & -9.2 & -0.1     \\
seg$_2$ & -0.1 & -0.3 & 0.5  &  ... & 0.3 & -0.5 & -0.9    \\
...   &      &      &      &  ...   &      &      &
\end{tabular}

\subsection{Test condition}


\begin{itemize}
\item No additional training materials, except AP17-OL3, AP16-OL7 and THCHS30, are allowed to be used.
\item All the trials should be processed. Scores of lost trials will be interpreted as -$\inf$.
\item The speech segments in each test condition (1 second, 3 second or the full-utterance)
      should be processed independently, and each test segment in a group should be processed
      independently too. Knowledge from other test segments is not allowed to use (e.g.,
      score distribution of all the test segments).
\item Information of speakers is not allowed to use.
\item Listening to any speech segments is not allowed.
\end{itemize}

\subsection{Evaluation metrics}

As in LRE15, the AP17-OLR challenge chooses $C_{avg}$ as the principle evaluation metric.
First define the pair-wise loss that composes the missing and
false alarm probabilities for a particular target/non-target language pair:

\[
C(L_t, L_n)=P_{Target} P_{Miss}(L_t) + (1-P_{Target}) P_{FA}(L_t, L_n)
\]

\noindent where $L_t$ and $L_n$ are the target and non-target languages, respectively; $P_{Miss}$ and
$P_{FA}$ are the missing and false alarm probabilities, respectively. $P_{target}$ is the prior
probability for the target language, which is set to $0.5$ in the evaluation. Then the principle metric
$C_{avg}$ is defined as the average of the above pair-wise performance:


\[
 C_{avg} = \frac{1}{N} \sum_{L_t} \left\{
\begin{aligned}
  & \ P_{Target} \cdot P_{Miss}(L_t) \\
  &  + \sum_{L_n}\ P_{Non-Target} \cdot P_{FA}(L_t, L_n)\
\end{aligned}
\right\}
\]

\noindent where $N$ is the number of languages, and $P_{Non-Target}$ = $(1-P_{Target}) / (N -1 )$.
We have provided the evaluation script for system development.

\section{Baseline systems}

\begin{figure}[hbt]
\begin{center}
\includegraphics[width=0.8\linewidth]{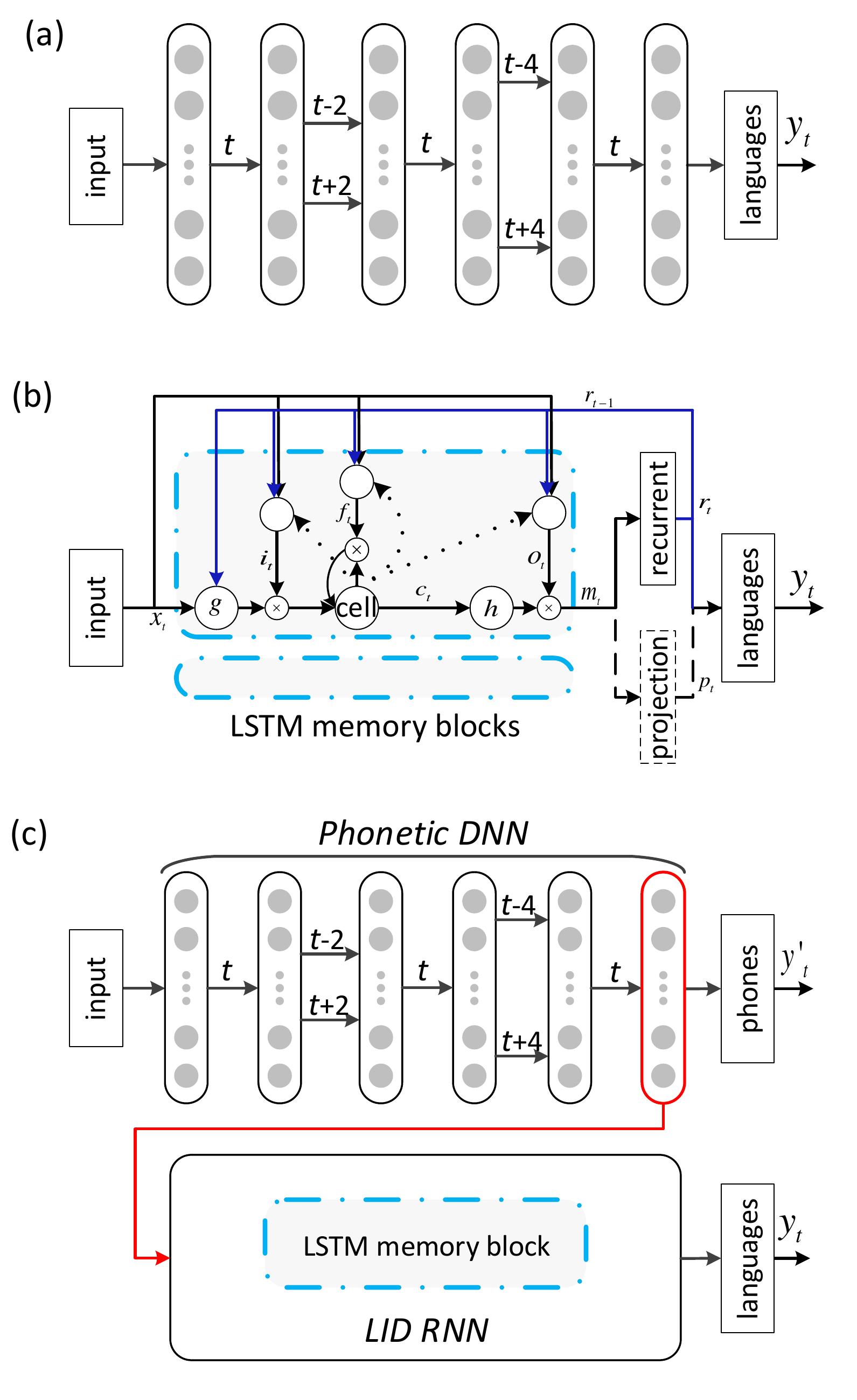}
\end{center}
\caption{Three neural-based LID architectures. The TDNN LID system (above), the LSTM-RNN LID system (middle) and
         the PTN LID system (bottom). In the PTN LID system, the phonetic feature is read from
         the last hidden layer of the phonetic DNN which is a TDNN and then becomes
         the only input for the system.}
\label{fig:lid}
\end{figure}

\begin{table*}[htb]
\normalsize
\begin{center}
\caption{C$_{avg}$ and EER results of various i-vector baseline systems on 3 test conditions.}
\label{tab:results1}
\begin{tabular}{|l|c|c|c|c|c|c|}
\hline
               & \multicolumn{2}{|c|}{1 second } & \multicolumn{2}{c|}{3 second }  & \multicolumn{2}{c|}{Full-Length}\\
\hline
System        &  $C_{avg}$  &   EER\%   & $C_{avg}$  &   EER\%  &  $C_{avg}$  &   EER\% \\
\hline
\hline
i-vector       & 0.1672 &	15.28           & 0.0695 &	7.59       & 0.0522 &	6.224\\
i-vector + LDA & 0.1238 &	13.30           & 0.0494 &	5.95       & 0.0362 &	4.704\\
\hline
\hline
i-vector  & 0.1485 &	14.43           & 0.0624 &	6.07       & 0.0469 &	4.58\\
(Linear SVM)      &                 &           &           &       &  &\\
i-vector  & 0.1242 &	12.43           & 0.0470 &	4.83       & 0.0351 &	3.58\\
(Poly SVM)        &                 &           &           &     & &\\
i-vector  & 0.1313 &	12.16           & 0.0495 &	4.59       & 0.0352 &	3.39\\
(RBF SVM)         &                 &           &           &     & &\\
\hline
\hline
L-vector + LDA  & 0.1336 &	12.47           & 0.0492 &	4.74       & 0.0337 &	3.30\\
(Linear SVM)      &                 &           &           &    & & \\
L-vector + LDA  & 0.1415 &	12.98           & 0.0538 &	4.92       & 0.0373 &	3.49\\
(Poly SVM)        &                 &           &           &    & & \\
L-vector + LDA  & 0.1370 &	12.61           & 0.0513 &	4.73      & 0.0355 &	3.32\\
(RBF SVM)         &                 &           &           &     & &\\
\hline
\end{tabular}
\end{center}
\end{table*}

\begin{table*}[htb]
\normalsize
\begin{center}
\caption{C$_{avg}$ and EER results of various DNN baseline systems on 3 test conditions.}
\label{tab:results2}
\begin{tabular}{|l|cc|cc|cc|cc|cc|cc|}
\hline
               & \multicolumn{4}{|c|}{1 second } & \multicolumn{4}{c|}{3 second }  & \multicolumn{4}{c|}{Full-Length}\\
\hline
    &  \multicolumn{2}{|c}{$C_{avg}$}  &\multicolumn{2}{c|}{EER\%}&\multicolumn{2}{|c}{$C_{avg}$}  &\multicolumn{2}{c|}{EER\%}&  \multicolumn{2}{|c|}{$C_{avg}$}  &\multicolumn{2}{c|}{EER\%} \\
\hline
System        & Fr. & Utt. & Fr. & Utt.                  & Fr. & Utt. & Fr. & Utt.               & Fr. & Utt. & Fr. & Utt. \\
\hline
\hline
TDNN-LID      & 0.1930 & 0.1604 &19.18 & 15.63           & 0.1920 & 0.1523 & 19.14 & 15.43       & 0.1925 & 0.1451 & 17.95 & 14.65\\
LSTM-LID      & 0.1837 & 0.1569 &19.13 & 16.77           & 0.1773 & 0.1525 & 18.90 & 16.99       & 0.1771 & 0.1468 & 17.87 & 16.03\\
PTN-LID       & 0.1821 & 0.1153 &18.43 & 11.88           & 0.1571 & 0.0727 & 16.03 & 8.24        & 0.1516 & 0.0689 & 14.82 & 8.15\\
\hline
\end{tabular}
\end{center}
\end{table*}

We constructed two kinds of baseline LID systems, based on the i-vector model and various DNN models respectively.
All the experiments were conducted with Kaldi~\cite{povey2011kaldi}.
The purpose of these experiments is not to present a competitive submission, instead
to present a reference for the participants. The recipes can be downloaded from the challenge web site.

\subsection{i-vector system}

The i-vector baseline systems were constructed based on the i-vector model~\cite{dehak2011front-end,dehak2011language}.
The static acoustic features involved 19-dimensional Mel
frequency cepstral coefficients (MFCCs) and the log energy.
This static features were augmented by their first and second order derivatives, resulting in 60-dimensional
feature vectors.
The UBM involved $2,048$ Gaussian components and the dimensionality of the i-vectors was $400$.
Linear discriminative analysis (LDA) was employed to promote language-related information.
The dimensionality of the LDA projection space was set to $6$.

With the i-vectors (either original or after LDA transform), the score of a trail on a particular language
can be simply computed as the cosine distance between the test i-vector and the mean i-vector of
the training segments that belong to that language. This is denoted to
be `cosine distance scoring'.
A more powerful scoring approach is to employ various discriminative models. In our experiment, we
trained a support vector machine (SVM)  for each language to determine the score
that a test i-vector belongs to that language. The SVMs were trained on the i-vectors
of all the training segments, following the one-verse-rest scheme. We will call this scoring
approach as `SVM-based scoring'.

\subsection{DNN systems}

For the DNN baseline, three kinds of DNN architectures were designed.
The first two are the traditional time-delay neural network (TDNN)~\cite{lang1990time} and
recurrent neural network with long short-term memory units (LSTM-RNN)~\cite{hochreiter1997long},
as shown in Fig.~\ref{fig:lid} (a) and (b) respectively.
The third one is based on the recently proposed phonetic temporal neural (PTN) model~\cite{tang2017phonetic},
where an auxiliary phonetic model produces phonetic feature, and an RNN LID model is used to
identify the language. The architecture is shown in Fig.~\ref{fig:lid} (c).
Both the the LSTM-RNN LID system and the PTN LID system employ an LSTM-RNN to identify
languages; the difference is that the PTN system uses a phonetic model to
extract phonetic features, rather than using raw acoustic features. Since the phonetic
model is trained with a different objective function (phone discrimination),
it can be seen as a mult-task training approach. In this study, the phonetic model
is a TDNN and was trained using the THCHS30 database and the accompanied Kaldi recipe.

The raw feature of all the three DNN systems is $40$-dimensional Fbanks,
with a symmetric $4$-frame window for the TDNN and
a symmetric $2$-frame window for the LSTM-RNN to splice neighboring frames.
For the TDNN LID, there are 6 hidden layers, and the activation function is p-norm. The
number of units of each TDNN layer is set to be $2048$.
The number of cells of the LSTM is set to be $1024$.

\subsection{Performance results}

The primary evaluation metric in AP17-OLR is $C_{avg}$. Besides that, we also present the performance
in terms of equal error rate (EER). These metrics evaluate
system performance from different perspectives, offering a whole picture of the verification/identification capability
of the tested system. The performance is evaluated on the development set at present.

The utterance level C$_{avg}$ and EER results of various i-vector baseline systems
are showed in Table~\ref{tab:results1}.
The frame and utterance level C$_{avg}$ and EER results of various DNN baseline systems
are showed in Table~\ref{tab:results2}.
From these results, we can observe that the i-vector systems generally perform well with long duration utterances.
On short utterances, the PTN system performs the best.

\section{Conclusions}

We presented the data profile and the evaluation plan of the AP17-OLR challenge.
To assist participants to establish a reasonable starting system, we published two types of baseline systems
based on the i-vector model and various DNN models respectively. All the data resources are free for
the participants, and the recipes of the baseline systems can be freely downloaded.


\bibliographystyle{IEEEtran}
\bibliography{ole}

\end{document}